\documentclass{svproc}

\usepackage{type1cm}        % activate if the above 3 fonts are
\usepackage{makeidx}         % allows index generation
\usepackage{graphicx}        % standard LaTeX graphics tool
\usepackage{multicol}        % used for the two-column index
\usepackage[bottom]{footmisc}% places footnotes at page bottom
\usepackage{caption}

\usepackage{array}
\usepackage{newtxtext}       % 
\usepackage[varvw]{newtxmath}       % selects Times Roman as basic font
\usepackage{xcolor} %which provides support for color manipulation \textcolor{}{}

\makeindex             % used for the subject index
\begin{document}
\title{Kinetostatic Analysis for 6RUS Parallel Continuum Robot using Cosserat Rod Theory}

%% svproc class
\author{Vinayvivian Rodrigues\inst{1,2} \and Bingbin Yu\inst{2} \and Christoph Stoeffler\inst{2} \and Shivesh Kumar\inst{2,3}}
\authorrunning{V. Rodrigues et al.} % abbreviated author list (for running head)
\tocauthor{Vinayvivian Rodrigues\inst{1,2}, Bingbin Yu\inst{2}, Shivesh Kumar\inst{2,3}, Christoph Stoeffler\inst{2}}
\institute{
Rheinisch-Westf{\"a}lische Technische Hochschule Aachen, 52062 Aachen, Germany\\
\email{vinayvivian.rodrigues@rwth-aachen.de}
\and
Robotics Innovation Center, German Research Center for Artificial Intelligence (DFKI GmbH), 28359 Bremen, Germany\\ 
\email{bingbin.yu@dfki.de, christoph.stoeffler@dfki.de}
\and
Department of Mechanics \& Maritime Sciences, Chalmers University of Technology, \\ 41296 G{\"o}teborg, Sweden
\email{shivesh@chalmers.se}
}
\maketitle

\abstract{Parallel Continuum Robots (PCR) are closed-loop mechanisms but use elastic kinematic links connected in parallel between the end-effector (EE) and the base platform. 
PCRs are actuated primarily through large deflections of the interconnected elastic links unlike by rigid joints in rigid parallel mechanisms.
In this paper, Cosserat rod theory-based forward and inverse kinetostatic models of 6-\underline{R}US PCR are proposed.
A set of simulations are performed to analyze the proposed PCR structure which includes maneuverability in 3-dimensional space through trajectory following, deformation effects due to the planar rotation of the EE platform, and axial stiffness evaluation at the EE.}

%%%%%%%%%%%%%%%%%%%%%%%%%%%%%%%%%%%%%%%%%%%%%
\section{Introduction}
\label{sec:1}
Parallel Kinematic Mechanisms (PKM) have received attention due to their higher stiffness, agility, and increased payload capacity when compared to the serial manipulators. They find applications in manufacturing, rehabilitation, medical surgery, vehicle simulators, and space docking systems. In many of these scenarios, compliance modulation of PKMs is desirable which can be introduced either via only the virtue of active control~\cite{DUTTA2019} or by introducing intrinsic physical compliance. The latter can be achieved by introducing springs in the joint space of the PKM (known as variable stiffness mechanisms, see~\cite{stoeffler2021comparative} for a comparative study) or by using flexible links in parallel assembly (parallel continuum robots~\cite{Black2014}). 

Cosserat rod theory~\cite{antman2005} can be used to study PCRs where the flexible links are implemented as slender rods. Black et al in \cite{Black2014} studied a Stewart Gough type PCR by formulating the static forward and inverse kinematics problems as the solution to multiple Cosserat-rod models with coupled boundary conditions. Till et al~\cite{John2015} improved the earlier work in \cite{Black2014} by introducing an efficient way to compute the error Jacobian. Later, Black et al~\cite{Black2018} further introduced a generalized framework for various kinematic configurations.

\begin{figure}[!ht]
\centering
\includegraphics[width=0.6\textwidth]{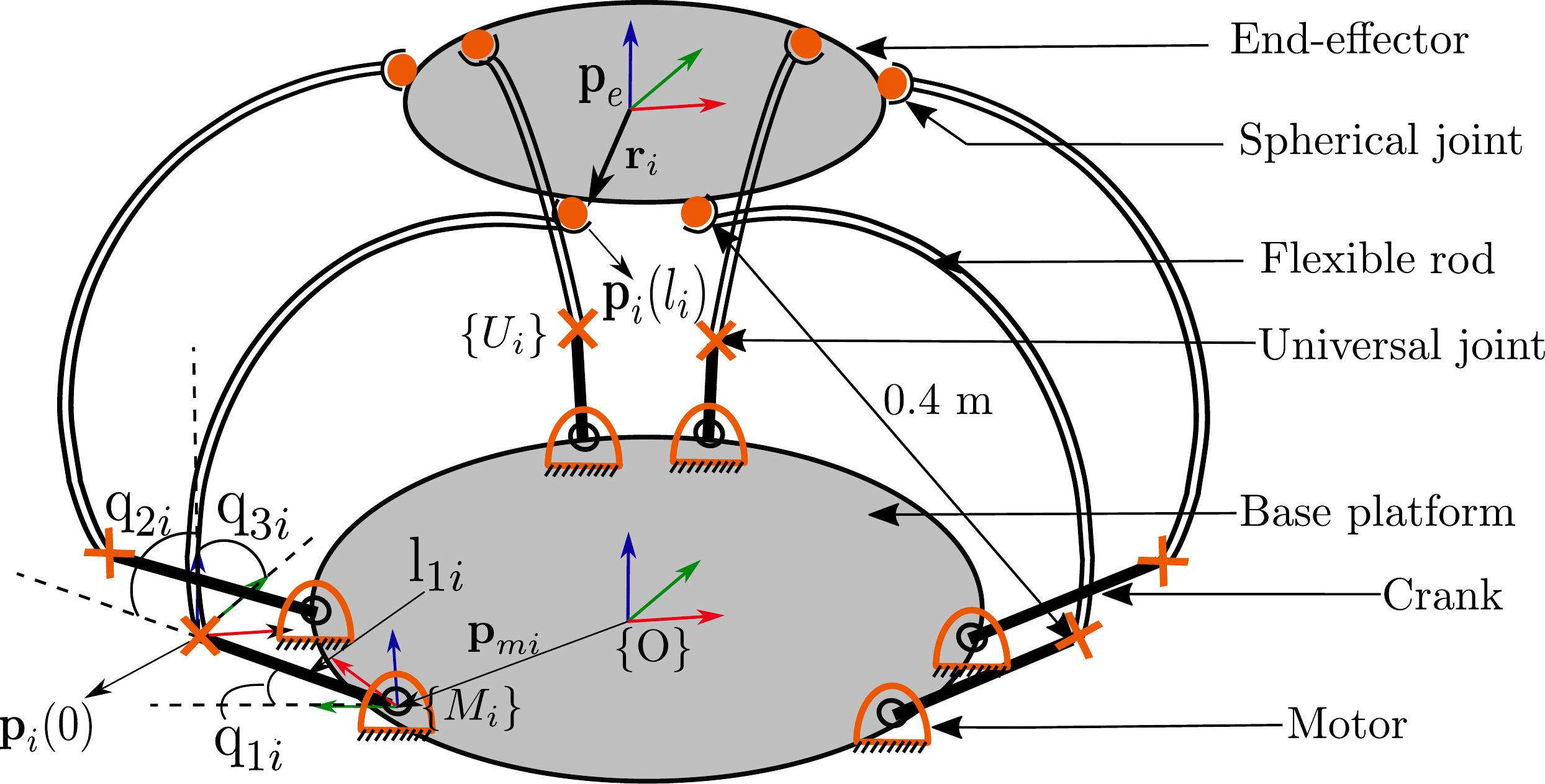}
\caption{Schematic explaining the kinematics for the proposed 6-\underline{R}US PCR.}
\label{cad}       % Give a unique label
\end{figure}

This paper presents the kinetostatic analysis of a novel 6-\underline{R}US PCR (see Fig.~\ref{cad}) using the Cosserat rod theory. We formulate the boundary conditions for solving the forward and inverse kinetostatics of the mechanism and provide position as well as stiffness analysis of the mechanism. We further provide insight into the range of motion through workspace evaluation for the PCR.

\paragraph{Organization:} The paper is organized as follows. Sec. \ref{math} presents the mathematical background on Cosserat rod theory relevant for modeling PCR based on~\cite{Black2018}. Sec. \ref{sec:BC} derives the boundary conditions for modeling forward and inverse kinetostatic model of 6-\underline{R}US PCR. Sec. \ref{sec:results} discusses the results including the workspace analysis, trajectory following, axial stiffness evaluation, etc. Sec. \ref{conclusion} concludes the paper and highlights the future work.

\section{Mathematical Background}
\label{math}
Consider a robot with $n$ elastic rods connected in parallel between the EE and the base platform. 
The shape of the $i^{th}$ rod is defined by its 
spatial centerline position, denoted by a function as \textbf{p}$_i$($s_i$) $\in$ $\mathbb{R}^3$. 
The centerline represents the tangent direction of the rod. The rod can have bending in $x$ and $y$ axes as well as torsion about the tangent $z$-axis. Thus, the rod state includes 3-DOF orientation, such that \textbf{R}$_i$($s_i$) $\in$ $SO(3)$. These functions are dependent on an arc length parameter, $s_i$ $\in$ [0, l$_i$], which represents the length 
of the rod along the centerline. $l_i$ is the total length of the $i^{th}$ rod. As the rod can have extension along the length, both $s$ and $l$ are defined in a no-stress configuration. Consequently, a homogeneous transformation, forming a material-attached frame to comprehensively describe the entire rod can be given as
\({\tiny
\textbf{H}_i(s_i) = \begin{bmatrix}\textbf{R}_i(s_i) & \textbf{p}_i(s_i) \\ 0 & 1\end{bmatrix}}\)$ \in SE(3)$.
\begin{figure}[!ht]
\centering
\includegraphics[width=0.6\textwidth]{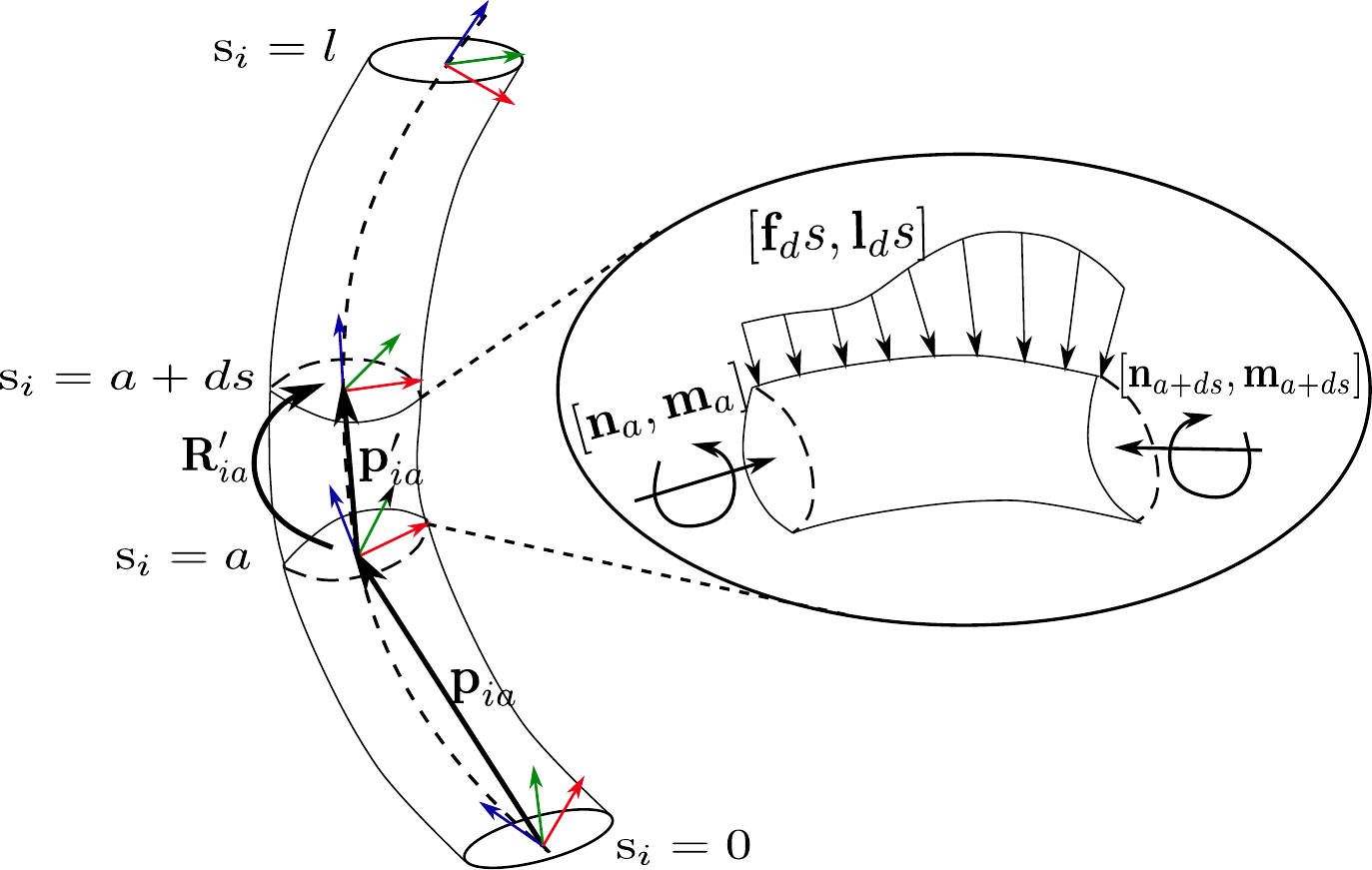}
\caption{Internal forces $\textbf{n} \in \mathbb{R}^3$ and internal moments $\textbf{m} \in \mathbb{R}^3$ acting on an arbitrary section of the rod. These forces and moments are expressed in the global frame H(s).}
\label{internal}       % Give a unique label
\end{figure}
The change in the position and orientation along the length of the elastic rod is defined by the linear rate vector \textbf{v}$_i(s)$ $\in$ $\mathbb{R}^3$,
and angular rate vector \textbf{u}$_i(s)$ $\in$ $\mathbb{R}^3$ where both are expressed in the 
local body coordinate system i.e. material frame which is given as:
\begin{equation}
\label{met:eq3}
\begin{aligned}
    \textbf{p}_i'(s_i) = \textbf{R}_i(s_i) \textbf{v}_i(s_i)\\
    \textbf{R}_i'(s_i) = \textbf{R}_i(s_i) \hat{\textbf{u}}_i(s_i)
\end{aligned}
\end{equation}
Where $(.)'$ denotes the derivative with respect to $s_i$ and $(.)\hat{}$ operator denotes mapping from $\mathbb{R}^3$ to $\mathfrak{so}(3)$ (the Lie algebra of SO(3)).
% \(\small
% \hat{\textbf{k}} = \begin{bmatrix}
%     0 & -k_{3} & k_{2} \\
%     k_{3} & 0 & -k_{1} \\
%     -k_{2} & k_{1} & 0  \\
% \end{bmatrix}\). 
Inverse mapping of $(.)\hat{}$ can be achieved as $\hat {(\textbf{u})}^V$ = $\textbf{u}$ 
where $(.)^V$ maps from $\mathfrak{so}(3)$ to $\mathbb{R}^3$.
Therefore, if one knows linear strain vector $\textbf{v}_i$ and angular strain vector $\textbf{u}_i$ in 
the material frame and an initial reference frame $\textbf{H}_i(s_i=0)$, the remaining frames can be obtained by integrating the differential equations mentioned in Eq. (\ref{met:eq3}). Since $\textbf{v}_i$ and $\textbf{u}_i$ changes 
with respect to $s_i$, numerical integration is needed to obtain $\textbf{H}_i(s_i)$.\par
The formulation for the classical Cosserat rod equations for static equilibrium is utilized from \cite{Black2018} and \cite{antman2005} which can be written as:
\begin{equation}
\label{force_eq}
\begin{gathered}
\textbf{n}_i^{'} = -\textbf{f}_i \\
\textbf{m}_i^{'} = -\textbf{p}_i^{'} \times \textbf{n}_i - \textbf{l}_i
\end{gathered}
\end{equation}
where Eq. (\ref{force_eq}) tracks the change in the $\textbf{n}$(s) and $\textbf{m}$(s) with respect to
the parameter $s$ along the length of the rod, and $\textbf{f}_i$ and $\textbf{l}_i$ are the distributed forces and moments per unit length of the rod respectively acting along the rod $i$ and are expressed in the world coordinates. Any external distributed force i.e. self weight is represented by $\textbf{f}_i$ as shown in Fig. (\ref{internal}). The rod self-weight can be written as \(\textbf{f} = \rho A \textbf{g}\) where $\rho$ is the density of the material, $A$ is the area of the cross-section, and $\textbf{g}$ is gravity constant vector acting along the $z$-axis. $\textbf{l}_i$ is assumed to be negligible in this work.\par 

One needs to combine the Cosserat rod equations in Eq. (\ref{force_eq}) with a material
constitutive model in order to connect displacements in the rod to the internal forces and moments. Throughout this work, a linear material constitutive relationship is used as
\(\small
    \textbf{n}_i(s) = \textbf{R}_i(s) \textbf{K}_{SE,i} (\textbf{v}_i(s) - \textbf{v}_i^*(s))\) and
    \(
    \textbf{m}_i(s) = \textbf{R}_i(s) \textbf{K}_{BT,i} (\textbf{u}_i(s) - \textbf{u}_i^*(s))
\)

where 
%\small
$\textbf{K}_{SE,i}(s)=\text{diag}(G_iA_i(s),G_iA_i(s),E_iA_i(s))$
% \) 
and
$\textbf{K}_{BE,i}(s)=\text{diag}(\\EI_{xx,i}(s),EI_{yy,i}(s),GJ)$.

Upon the external load at any given point $s_i$ on the elastic rod, the diagonal matrices $\textbf{K}_{SE,i}(s)$ and $\textbf{K}_{BT,i}(s)$ depict the mechanical stiffness for unit length. The subscript “SE” refers to shear and extension, and “BT” refers to bending and torsion. Whereas $A$ is the cross-sectional area, $E$ is Young’s modulus, $G$ is Shear modulus, $I_{xx}$ and $I_{xx}$ are the second area moment (about the local
$x$ and $y$ axes), and $J$ is the polar area moment about the local $z$-axis. $\textbf{v}_i^*(s)$ and $\textbf{u}_i^*(s)$ represents the reference strain vectors at the rest configuration. The pre-curved configuration is expressed as $\textbf{u}_i^*(s)=[1/r,0,0]^T$ where $r$ is the radius of bent curvature with a value of 0.3005 m and for a straight rod, $\textbf{v}_i^*(s)=[0,0,1]^T$.\par
Thus Eqs. (\ref{met:eq3}) and  (\ref{force_eq}) form a system of 
differential equations that describe the rate of change of the state variables $\textbf{p}_i$, $\textbf{R}_i$,
$\textbf{n}_i$, $\textbf{m}_i$ with respect to the arc parameter $s_i$.

%%%%%%%%%%%%%%%%%%%%%%%%%%%%%%%%%%%%%%%%%%%%%%%%%%%%

\section{Kinetostatic Model of 6-RUS PCR}
\label{sec:BC}
% At the distal end of the 6-\underline{R}US PCR, conditions of static equilibrium are considered for the 
% end-effector. 
Due to the closed-loop parallel kinematics, the physical constraints are coupled with the boundary conditions for each independent elastic rod irrespective of joint type present in the PCR.
At the distal end i.e $s_i = l_i$, for each elastic rod the following static equilibrium equations hold:
\begin{equation}
\label{force_bal}
\begin{aligned}
    &\sum_{i=1}^{6} [\textbf{n}_i(l_i)] - \textbf{F} = 0\\
    &\sum_{i=1}^{6} [\textbf{p}_i(l_i)\times\textbf{n}_i(l_i)+\textbf{m}_i(l_i)] - \textbf{p}_e\times \textbf{F} - \textbf{M} = 0
    \end{aligned}
\end{equation}
Here, $\textbf{F}\in \mathbb{R}^3$ and $\textbf{M}\in \mathbb{R}^3$ are the external force and moment, respectively acting at the center of the EE, $\textbf{p}_e \in \mathbb{R}^3$ which are expressed in the global coordinates. 
\subsection{Boundary Conditions}
\paragraph{Distal joint boundary condition ($s_i=l_i$)}
Eq.(\ref{pos_error}) constraints the end tip position $\textbf{p}_i(l_i)$ of each elastic rod to the EE platform. $\textbf{r}_i$ is the constant vector which represents the position of the elastic rod tip in EE coordinate frame as seen in Fig. (\ref{cad}). 
\begin{equation}
\label{pos_error}
    \textbf{p}_e + \textbf{R}_e \textbf{r}_i - \textbf{p}_i(l_i) = 0 \quad\text{for i = 1,...,6}
\end{equation}
Due to the spherical joint at the tip of each link, there are no restriction on the orientation of the material as the spherical joint cannot transmit moments, which results in constraints expressed by Eq. (\ref{moment_error}).
\begin{equation}
\label{moment_error}
  \textbf{R}_e^T\textbf{m}_i(l_i) = 0\quad\text{for i = 1,...,6}
\end{equation}
Here $\textbf{R}_e(\alpha, \beta, \gamma) \in \mathbb{R}^{3 \times 3}$ is the rotation matrix of the EE  in global coordinates parameterized by Roll$(\alpha)$--Pitch$(\beta)$--Yaw$(\gamma)$ angles. 
Residual vector $\textbf{E} \in \mathbb{R}^{42}$ describes the total error for a solution that should be zero which is expressed as 
    \(\textbf{E}=[[(\textbf{E}^{\textbf{F}})^T,(\textbf{E}^{\textbf{M}})^T]^T,(\textbf{E}_{1}^{\textbf{p}_{01}})^T,\dots,(\textbf{E}_{6}^{\textbf{p}_{06}})^T, (\textbf{E}_{1}^{\textbf{m}_{1}})^T,\dots, (\textbf{E}_{6}^{\textbf{m}_{6}})]^T\)
where terms $\textbf{E}^{\textbf{F}}$ and $\textbf{E}^{\textbf{M}}$, $\textbf{E}_{i}^{\textbf{p}_{i}}$, $\textbf{E}_{i}^{\textbf{m}_{i}}$ are the error terms corresponding to Eqs.(\ref{force_bal}), (\ref{pos_error}), and (\ref{moment_error}) respectively.  
\par

\paragraph{Proximal joint boundary condition ($s_i=0$):}
At the base of the elastic rod, each link is connected to a rigid crank with a universal joint as shown in Fig. (\ref{cad}). The pose, $\textbf{p}_{0i} \in \mathbb{R}^3$ and $\textbf{R}_{0i} \in SO(3)$ of the universal joint in the world frame \{$O$\} can be expressed as: $\textbf{p}_{0i}=\textbf{p}_{mi}+\textbf{R}_{M_i}^O(q_{1i})\textbf{p}^{M_i}_{0i}$ where $\textbf{p}^{M_i}_{0i}$ is the position of the universal joint in motor frame \{$M_i$\} given by $[0, l_{1i} \cos{(q_{1i})}, l_{1i} \sin{(q_{1i})})]$ where $l_{1i}$ is the length of the crank and $q_{1i}$ is the motor angle made with the $y$-axis of the \{$M_i$\}. $\textbf{p}_{mi}$ is the constant vector that describes the position of the motors in \{$O$\} and $\textbf{R}_{M_i}^O(q_{1i}) \in SO(3)$ is the orientation of \{$M_i$\} in \{$O$\}. Whereas, $\textbf{R}_{0i}=\textbf{R}_{M_i}^O(q_{1i})\textbf{R}^{M_i}_{U_i}(q_{2i},q_{3i})$ in which $\textbf{R}^{M_i}_{U_i}(q_{2i},q_{3i})=\textbf{R}^{M_i}_{U_{2i}}(q_{2i})\textbf{R}^{U_{2i}}_{U_{1i}}(q_{3i})$ where $q_{2i}$ and $q_{3i}$ are the universal joint angles about $x$ and $y$ axes in the universal joint frame \{$U_i$\}, $\textbf{R}^{M_i}_{U_i}$ describes the orientation of \{$U_i$\} in \{$M_i$\}, and $\textbf{R}^{U_{2i}}_{U_{1i}}$ denotes the local orientation between the individual revolute joints in \{$U_i$\}.

Due to the universal joint at $s_i=0$, it allows zero moment about $x$ and $y$ axes but transmits non-zero moment about the local $z$-axis. Then the moments at $s_i = 0$ can be set as: $m_{ix}(0) = m_{iy}(0) = 0$ while $m_{iz}(0)$ will be unknown. Regarding the forces at $s_i=0$, the shear forces $n_{ix}(0)$,$n_{iy}(0)$ and axial force, $n_{iz}(0)$ will be unknowns as well. Unknown vector $\textbf{B}_i$ describes the set of unknowns for each rod whose values will be guessed while integrating Eqs. (\ref{met:eq3}) and (\ref{force_eq}) till the tip of the rod as an initial value problem (IVP). These four unknowns are independent of the kinetostatic problem (IK and FK).

\subsection{Inverse and Forward Kinetostatic formualtion}
\paragraph{Inverse Kinetostatic (IK) model:}
IK model can be described as $\textbf{q}_1=IK(\textbf{p}_e,\textbf{R}_e,\\\textbf{F},\textbf{M},\textbf{B}_{IK},\textbf{H}_0)$ where vector $\textbf{q}_1$ consists of actuator variables and $\textbf{B}_{IK}$ contains the unknown variables for the system for the IK model.
In IK, additionally $q_{1i}$, $q_{2i}$, and $q_{3i}$ is included as unknown variables for each rod such that $\textbf{B}_{IK} \in \mathbb{R}^{42}$, expressed as 
\(\textbf{B}_{IK}=[n_{x,1}(0), n_{y,1}(0), n_{z,1}(0),m_{z,1}(0),q_{11}, q_{21}, q_{31},\dots]^T\).
\paragraph{Forward Kinetostatic (FK) model:}
Similarly, for the FK model can be described as function $\textbf{p}_e, \textbf{R}_e =FK(\textbf{q}_1,\textbf{F},\textbf{M},\textbf{B}_{FK},\textbf{H}_0)$) where $\textbf{B}_{FK}$ contains the unknown variables for the system for the FK model. In this case, $\textbf{p}_e$, $\textbf{R}_e$, $q_{2i}$, and $q_{3i}$ is considered as unknown variables such that $\textbf{B}_{FK} \in \mathbb{R}^{42}$ for the coupled system is given by:
\(\textbf{B}_{FK}=[n_{x,1}(0), n_{y,1}(0), n_{z,1}(0),m_{z,1}(0),q_{21}, q_{31},\dots,\textbf{p}_e, \textbf{R}_e]^T\).
\begin{figure}[!ht]
\centering
\includegraphics[width=\textwidth]{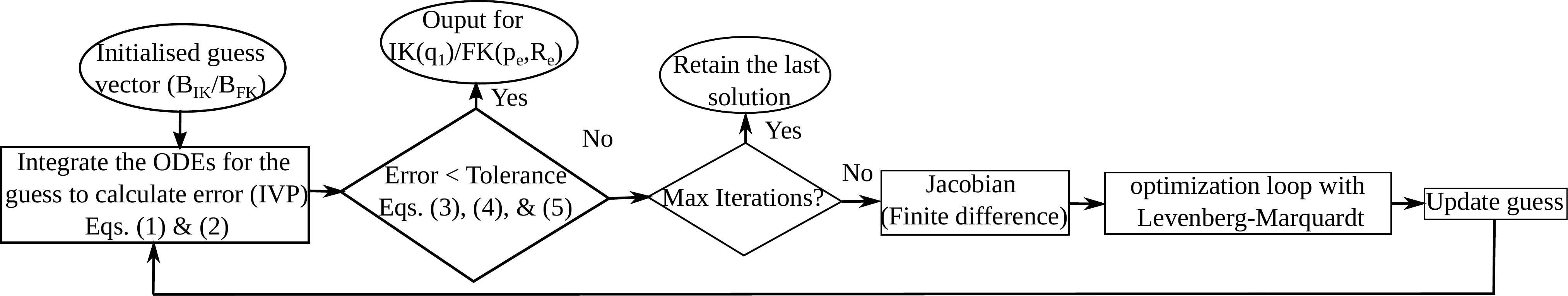}
\caption{A shooting method overview to evaluate the boundary value problem~\cite{John2015}.}
\label{flow}       % Give a unique label
\end{figure}
\subsection{Shooting Method Implementation}
% \paragraph{Shooting Method Implementation:}
From the earlier section, $\textbf{B}$ and $\textbf{E}$, forms a $42\times42$ system of equations for both IK and FK model. These equations are solved by using a shooting method as outlined in Fig. (\ref{flow}) to iteratively solve the coupled system for the vector $\textbf{B}_i$. In this approach, $\textbf{B}_i$ is initialized through a guessed set of values, and an initial value problem (IVP) for individual elastic rods, described in Eqs. (\ref{met:eq3}) and (\ref{force_eq}), is integrated till the tip of the link. Scipy library, \verb|integrate.solve_ivp|
function is used to numerically integrate the system of ODEs using an explicit Runge-Kutta method of order 5 for a given initial value. The residual vector $\textbf{E}_i$ is then computed for each independent IVP, and the unknown vector $\textbf{B}_i$ is iteratively updated by computing the Jacobian, \(
J = \left[ \frac{\partial \textbf{E}_i}{\partial \textbf{B}_i} \right]\)
via finite differences until the residual vector within a defined tolerance is achieved. For the non-linear optimization loop to update the unknowns, Scipy library \verb|optimize.root| is used which solves the system of nonlinear equations in the least squares sense with Levenberg-Marquardt algorithm.
%%%%%%%%%%%%%%%%%%%%%%%%%%%%%%%%%%%%%%%%%%%%%%%%%%%%
\section{Results}
\label{sec:results}
To validate the boundary condition formulation described in Sec. \ref{sec:BC}, titanium alloy (Ti6AL-4V) material properties are considered for the elastic rods. The rod is 4 mm in diameter and 530 mm long for which the Young’s modulus is 110.3 GPa, Poisson’s ratio is 0.31,  and density is 4428.8 Kg/m$^3$. As shown in Fig. \ref{cad}, the rod is connected to a crank at the base through a universal joint, and the tip of the rod is connected to the EE platform via a spherical joint. The opposite end of the crank is attached to the motor at the base. Also, the PCR has a pre-curved rest configuration as shown in Fig. \ref{cad}, such that the length is then reduced to 400 mm.\par
For the results obtained from the kinetostatic model, a tolerance of $5\times10^{-10}$ units is considered for each term in the residual vector $\textbf{E}$. Also a constraint of $\geq -20$ degrees till $\leq 90$ degrees is considered for the motor angles. The software implementation has been made available open-source~\footnote{\url{https://github.com/dfki-ric-underactuated-lab/6rus\_cosserat\_kinetostatics}}. \par
\begin{figure}[!ht]
\centering
\includegraphics[width=0.65\textwidth]{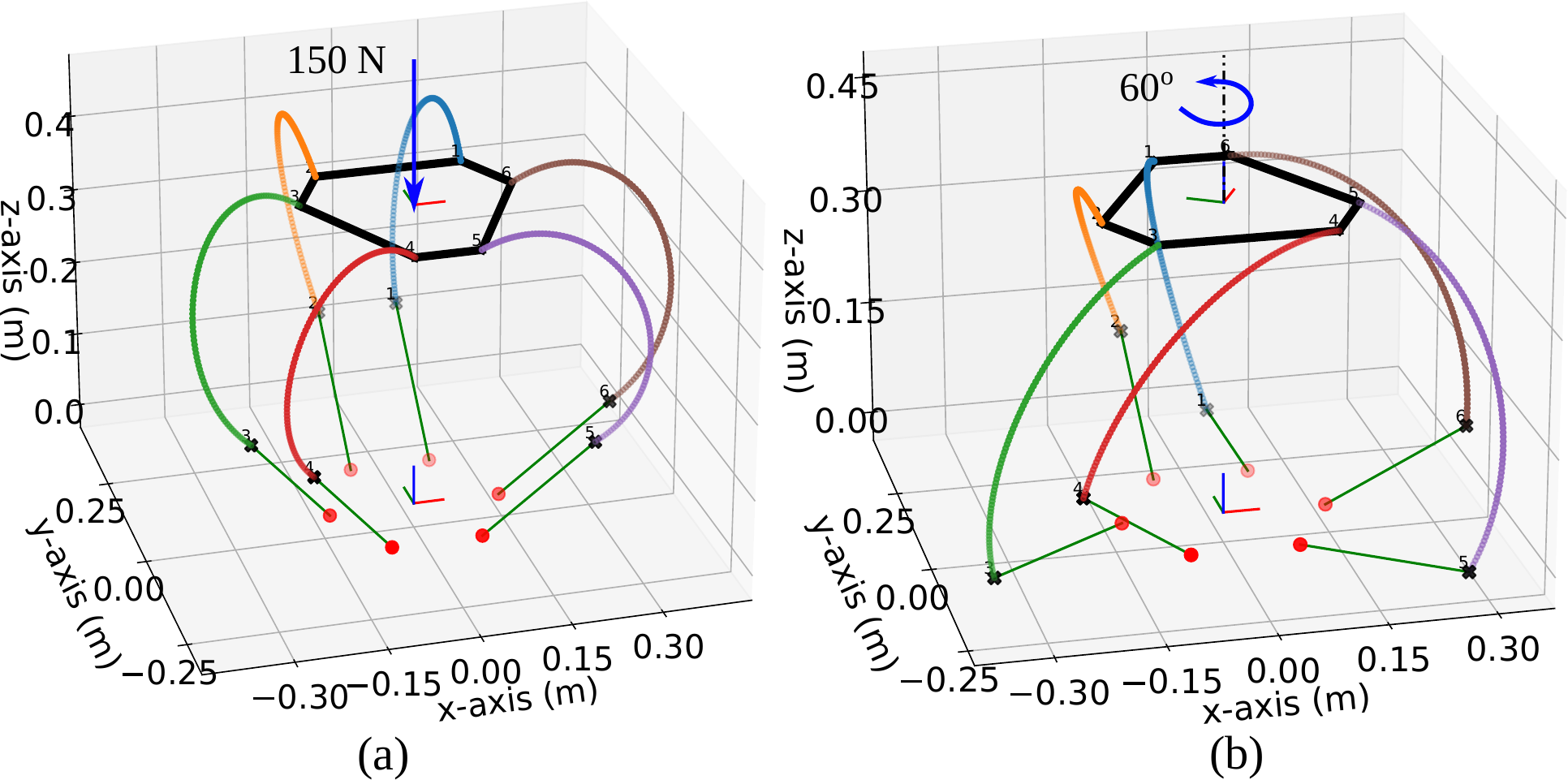}
\caption{(a) IK solution describing the deformation with 150 N acting at the EE. (b) IK solution for a rotated EE platform by $+60^\circ$ about the $z$-axis.}
\label{stiffness}       % Give a unique label
\end{figure}

\paragraph{Analysis for compressing forces at the EE:}
To evaluate the axial force response at the EE, IK model is used to solve for the external compressing forces sampled between 10 N and 300 N that is applied at the center of the EE at the rest height of 0.4 m. For the defined tolerance, a maximum compressing force is estimated to be 267 N for the PCR system. For every applied force, the corresponding change in the height of the EE in the world frame is also recorded for which the axial stiffness is estimated to be 989.75 N/m for the proposed PCR. The axial force response is illustrated in Fig. \ref{stiffness}a. \par
\paragraph{Analysis of rotation range of the EE:}
Another simulation is carried out to evaluate the solution for different rotation of the EE platform. IK model is used to solve for different angles about $z$-axis for the given EE pose. Here EE mass is considered negligible for the simplicity of the simulation. Note this simulation does not evaluate the twist angle of the individual elastic rod. The PCR is initialized at a rest EE height of 0.4 m where IK solution is recorded for yaw angle sampled between 0$^\circ$ to 90$^\circ$ in the positive (anti-clockwise) direction about the z-axis. Fig. \ref{stiffness}b, demonstrates the deformation caused by rotation of the EE platform.

\begin{figure}[!ht]
\centering
\includegraphics[width=0.68\textwidth]{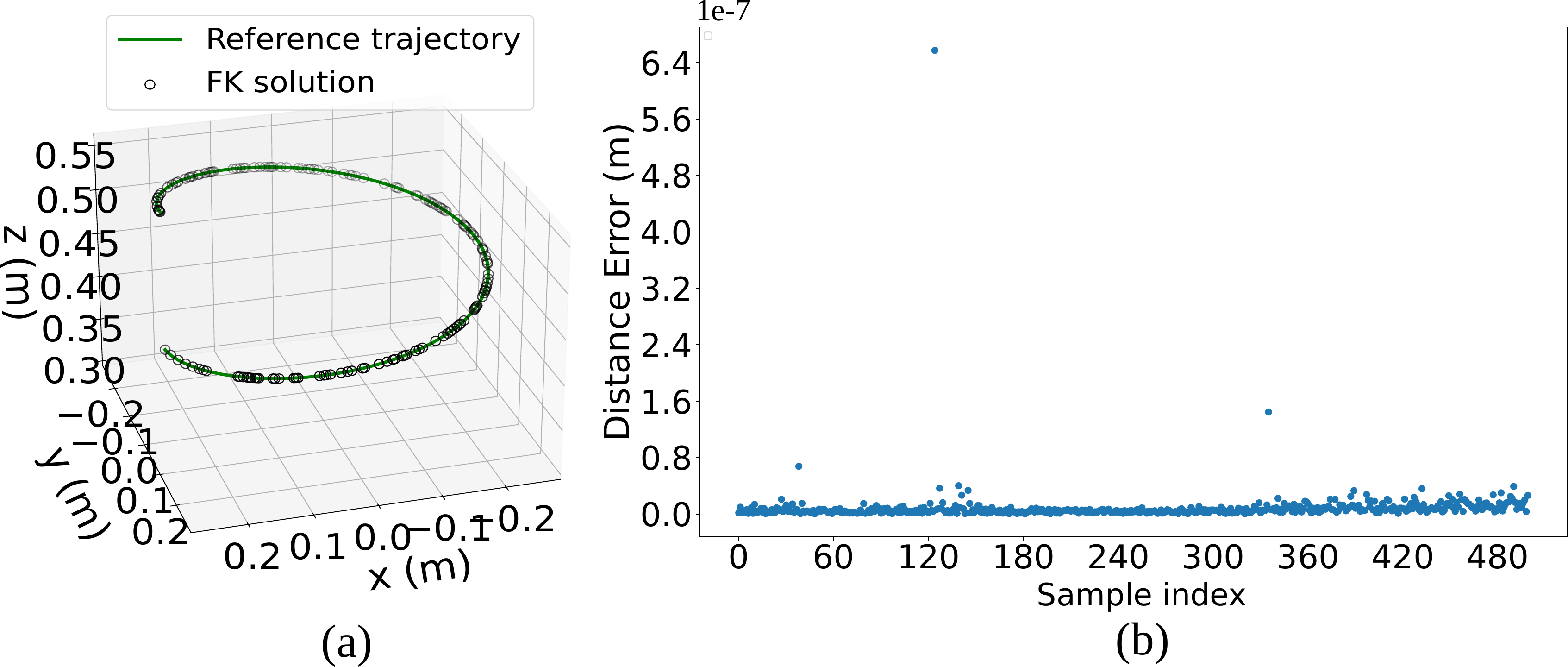}
\caption{(a) Helical trajectory comparison under 5 N load. (b) Euclidean distance between the reference EE position and the FK solution.}
\label{trajec}       % Give a unique label
\end{figure}

\paragraph{Trajectory following of the EE under constant load:}
In this simulation, the FK model is validated by comparing the obtained solution of the EE position with samples from a reference helical trajectory under a constant load of 5 N at the EE which is depicted in Fig. \ref{trajec}a. Euclidean distance is calculated to measure the error between the FK model and the reference trajectory. As seen in Fig. \ref{trajec}b, the error is of the order $1\times10^{-7}$ for the samples which shows the validity of the boundary conditions for the FK model for the PCR.

\begin{figure}
\centering
\includegraphics[width=0.65\textwidth]{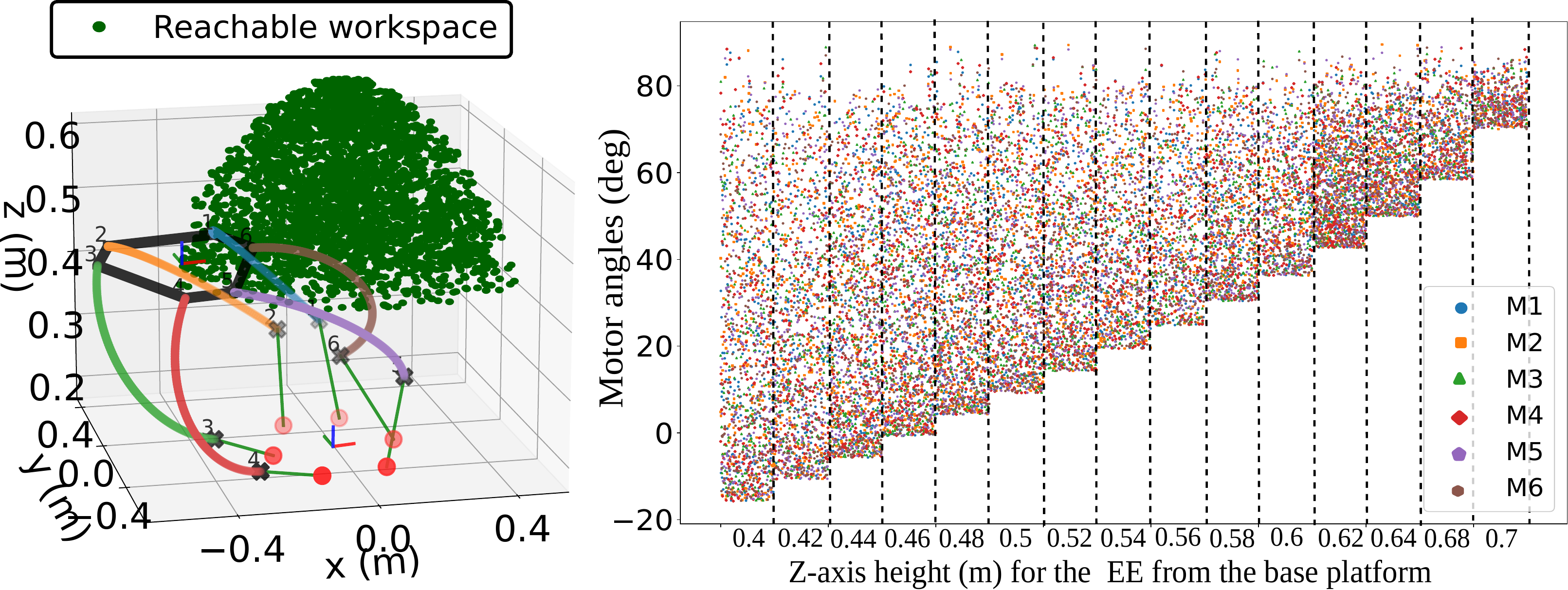}
\caption{(a) Reachable workspace in green for the PCR. (b) Range of motor angles for samples with increasing EE height.}
\label{workspace} 
\end{figure}
\par
\paragraph{Workspace analysis for the PCR:}
In this section, a reachable workspace is estimated for a 6-\underline{R}US PCR. Estimating a workspace means that finding solutions for the boundary value problem within the tolerance for the
boundary conditions defined for the PCR. Due to redundancy in the elastic rod, just providing the joint angles will give a different solution. The computational time taken by the boundary value problem is also large
as it needs to estimate the pose of the end-effector. So IK model is used to find a solution for heuristically provided sampled points from a cylindrical volume which represents the EE position in Cartesian space. For each sample point, IK solution is calculated, and then based on the boundary condition tolerances, only valid solutions are considered as part of the reachable workspace which is illustrated in Fig. \ref{workspace}a. Each partitions shown in Fig. \ref{workspace}b describes the motor angle range for the samples at a particular EE height. It is evident that the motor angle values are limited along the height due to the kinematic constraint of the system. For total 4000 samples, the mean computational time for the workspace samples is estimated to be 3.82 seconds with standard deviation of 1.15 seconds.
\section{Conclusion}
\label{conclusion}
In this paper, boundary conditions for both IK and FK are formulated for a 6-RUS PCR using Cosserat rod theory. A shooting method is used to iteratively solve the IVP by updating the guessed variables till the boundary value constraints are within the desired tolerance. The kinetostatic model has been analysed on a different aspect in simulation. Trajectory simulation shows the FK was able to find a solution with an error of the order $1\times10^{-7}$ under constant load condition of 5 N for a helical trajectory. Maximum load capacity and axial stiffness is estimated for the PCR by applying compressing forces at the EE. The solution for different EE rotation is studied to evalute the range of motion for the PCR. A reachable workspace is estimated for the proposed PCR using the IK model. Motor angles range for each rod are also visualised for the reachable workspace. The future work includes experimental validation of this model on the physical prototype. 
\section*{Acknowledgement}
The work presented in this paper is supported by the PACOMA project (Grant No. ESA-TECMSM-SOW-022836) subcontracted to us by Airbus Defence \& Space GmbH (Grant No. D.4283.01.02.01) with funds from the European Space Agency. The authors also want to acknowledge John Till's GitHub tutorial\footnote{\url{https://github.com/JohnDTill/ContinuumRobotExamples}} on PCR and his guidance on deriving the boundary condition equations for the proposed PCR.
\bibliographystyle{ieeetr}
\bibliography{references}

\begin{thebibliography}{1}

\bibitem{DUTTA2019}
A.~Dutta, D.~H. Salunkhe, S.~Kumar, A.~D. Udai, and S.~V. Shah, ``Sensorless full body active compliance in a 6 dof parallel manipulator,'' {\em Robotics and Computer-Integrated Manufacturing}, vol.~59, pp.~278--290, 2019.

\bibitem{stoeffler2021comparative}
C.~Stoeffler, S.~Kumar, and A.~M{\"u}ller, ``A comparative study on 2-dof variable stiffness mechanisms,'' in {\em Advances in Robot Kinematics 2020}, pp.~259--267, Springer, 2021.

\bibitem{Black2014}
C.~E. Bryson and D.~C. Rucker, ``Toward parallel continuum manipulators,'' in {\em 2014 IEEE International Conference on Robotics and Automation (ICRA)}, pp.~778--785, 2014.

\bibitem{antman2005}
S.~S. Antman, ``Problems in nonlinear elasticity,'' {\em Nonlinear Problems of Elasticity}, pp.~513--584, 2005.

\bibitem{John2015}
J.~Till, C.~E. Bryson, S.~Chung, A.~Orekhov, and D.~C. Rucker, ``Efficient computation of multiple coupled cosserat rod models for real-time simulation and control of parallel continuum manipulators,'' in {\em 2015 IEEE International Conference on Robotics and Automation (ICRA)}, pp.~5067--5074, 2015.

\bibitem{Black2018}
C.~B. Black, J.~Till, and D.~C. Rucker, ``Parallel continuum robots: Modeling, analysis, and actuation-based force sensing,'' {\em IEEE Transactions on Robotics}, vol.~34, no.~1, pp.~29--47, 2018.

\end{thebibliography}

\end{document}